\pdfoutput=1
\documentclass[letterpaper, 10pt, journal, twoside]{IEEEtran}

\usepackage{graphicx}
\usepackage{amsmath,amssymb,amsfonts}
\usepackage{booktabs}
\usepackage{multirow}
\usepackage{multicol}
\usepackage{siunitx}
\usepackage{url}
\usepackage{adjustbox}
\usepackage{threeparttable}
\usepackage{graphicx}
\usepackage{subcaption}
\usepackage{comment}
\usepackage{cite}
\usepackage{pifont} 
\usepackage{makecell} 

\usepackage{xcolor}
\definecolor{darkspringgreen}{rgb}{0.09, 0.45, 0.27}
\definecolor{darkred}{rgb}{0.55, 0.0, 0.0}
\newcommand{\cmark}{\textcolor{darkspringgreen}{\ding{51}}}
\newcommand{\xmark}{\textcolor{darkred}{\ding{55}}}

\def\bx{\mathbf{x}}
\def\Rot{R}

\def\MYTITLE{On the Benefits of Visual Stabilization for Frame- and Event-based Perception}

\newif\ifclarification
\clarificationtrue
\newif\ifpreprint
\preprinttrue

\author{J.P. Rodr\'{\i}guez-G\'omez\ifpreprint\orcidlink{0000-0001-7628-1660}\fi,  
J.R. Mart\'{\i}nez-de Dios\ifpreprint\orcidlink{0000-0001-9431-7831}\fi, 
A. Ollero\ifpreprint\orcidlink{0000-0003-2155-2472}\fi,
G. Gallego\ifpreprint\orcidlink{0000-0002-2672-9241}\fi%
\ifclarification\vspace{-2.5ex}\fi%
\thanks{Manuscript received February 29, 2024; Revised May 17, 2024; Accepted August, 6, 2024.}
\thanks{This paper was recommended for publication by Editor Cesar Cadena Lerma upon evaluation of the Associate Editor and Reviewers’ comments}
\thanks{This work was funded by GRIFFIN ERC Advanced Grant (Action 788247) from the European Research Council.
Funded by the Deutsche Forschungsgemeinschaft (DFG, German Research Foundation) under Germany’s Excellence Strategy – EXC 2002/1 ``Science of Intelligence'' – proj.  num. 390523135.\\
\ifclarification
J.P. Rodr\'{\i}guez-G\'omez is with Leonardo Labs, Leonardo S.p.A, Italy. This research was conducted while the author was at the GRVC Lab. Leonardo S.p.A was not involved in the development of this article or its content.\\
J.R. Mart\'{\i}nez-de Dios and A. Ollero are with the GRVC Robotics Laboratory of the University of Seville, Seville, Spain.
\else
J.P. Rodr\'{\i}guez-G\'omez, J.R. Mart\'{\i}nez-de Dios and A. Ollero are with the GRVC Robotics Laboratory of the University of Seville, Seville, Spain.
\fi
\newline 
G. Gallego is with the TU Berlin, the Einstein Center Digital Future, 
the SCIoI Excellence Cluster, and the Robotics Institute Germany, Berlin, Germany.%
}%
\thanks{Digital Object Identifier (DOI): \href{https://doi.org/10.1109/LRA.2024.3450290}{10.1109/LRA.2024.3450290}}
}

\definecolor{light-gray}{gray}{0.7}
\newcommand\gframe[1]{{\color{light-gray}\frame{#1}}}

\definecolor{darkcerulean}{rgb}{0.03, 0.27, 0.49}
\usepackage[pagebackref=false,breaklinks=true,colorlinks,allcolors=darkcerulean,bookmarks=true,bookmarksnumbered=true]{hyperref}
\hypersetup{
  pdftitle={\MYTITLE},
  pdfsubject={Robotics, Computer Vision},
  pdfauthor={J.P. Rodriguez-Gomez,  J.R. Martinez-de Dios, A. Ollero, G. Gallego},
  pdfkeywords={Stabilization, Event Cameras, Motion Estimation, Asynchronous Sensor, High Dynamic Range, High Temporal Resolution}
}

\usepackage{orcidlink} 

\usepackage[capitalize]{cleveref}
\crefname{section}{Sec.}{Secs.}
\crefname{table}{Tab.}{Tabs.} 
\crefname{figure}{Fig.}{Figs.}
\Crefname{section}{Section}{Sections}
\Crefname{table}{Table}{Tables}
\Crefname{figure}{Figure}{Figures}

\newif\ifclearlook

\def\flyone{\emph{indoor\_flying1}}
\def\flytwo{\emph{indoor\_flying2}}
\def\flythree{\emph{indoor\_flying3}}
\def\sofanormal{\emph{sofa\_normal}}
\def\sofanormalimu{\emph{sofa\_normal\_imu}}
\def\desknormal{\emph{desk\_normal}}
\def\desknormalimu{\emph{desk\_normal\_imu}}
\def\mountainnormal{\emph{mnt\_normal}}
\def\mountainnormalimu{\emph{mnt\_normal\_imu}}
\def\angvel{\boldsymbol{\omega}} 
\def\linvel{\mathbf{V}}  
\def\flowvec{\mathbf{u}} 
\def\invdepth{\rho}
\def\invdepthvec{\boldsymbol{\rho}}
\def\systemMatrix{\mathtt{A}}
\def\Kint{\mathtt{K}} 
\def\Rot{\mathtt{R}} 
\def\mH{\mathtt{H}} 

\title{\MYTITLE}

\begin{document}
\maketitle

\begin{abstract}
Vision-based perception systems are typically exposed to large orientation changes in different robot applications. 
In such conditions, their performance might be compromised due to the inherent complexity of processing data captured under challenging motion.  
Integration of mechanical stabilizers to compensate for the camera rotation is not always possible due to the robot payload constraints.
This paper presents a processing-based stabilization approach to compensate the camera's rotational motion both on events and on frames (i.e., images). 
Assuming that the camera's attitude is available, we evaluate the benefits of stabilization in two perception applications: 
feature tracking and estimating the translation component of the camera's ego-motion. 
The validation is performed using synthetic data and sequences from well-known event-based vision datasets. 
The experiments unveil that stabilization can improve feature tracking and camera ego-motion estimation accuracy in 27.37\% and 34.82\%, respectively. 
Concurrently, stabilization can reduce the processing time of computing the camera's linear velocity by at least 25\%. Code is available at {\hypersetup{urlcolor=magenta}\href{https://github.com/tub-rip/visual_stabilization}{https://github.com/tub-rip/visual\_stabilization}}.
\end{abstract}


\section{Introduction}
\label{sec:intro}
\IEEEPARstart{F}{rom} the vibrations caused by the gait motion of humanoids to the intrinsic attitude changes produced by quadrotor's flight, vision sensors are often exposed to abrupt motion variations. 
These types of motions can compromise the quality of the sensing data and the performance of perception algorithms, which affect the decision and control functionalities of robots. Stabilization becomes a handy tool to reduce the complexity of processing data captured under linear and angular motions, and helps to improve the performance of the subsequent algorithms in the perception pipeline. 

In nature, birds use their neck to perform head stabilization while flying and hunting \cite{ros2017frontneurocience}. 
Humans regularly compensate their gaze to provide stable images on the retina during active movements \cite{crane1997jneurophysiology}. 
Similarly, mechanical stabilizers (i.e., gimbals) are installed in robotic platforms to compensate for camera rotations. 
However, their integration on lightweight robots is often limited due to the reduced payload of the platforms \cite{RodriguezGomez22ral}.  

\begin{figure}[t!]
\centering
    \includegraphics[width=.63\linewidth]{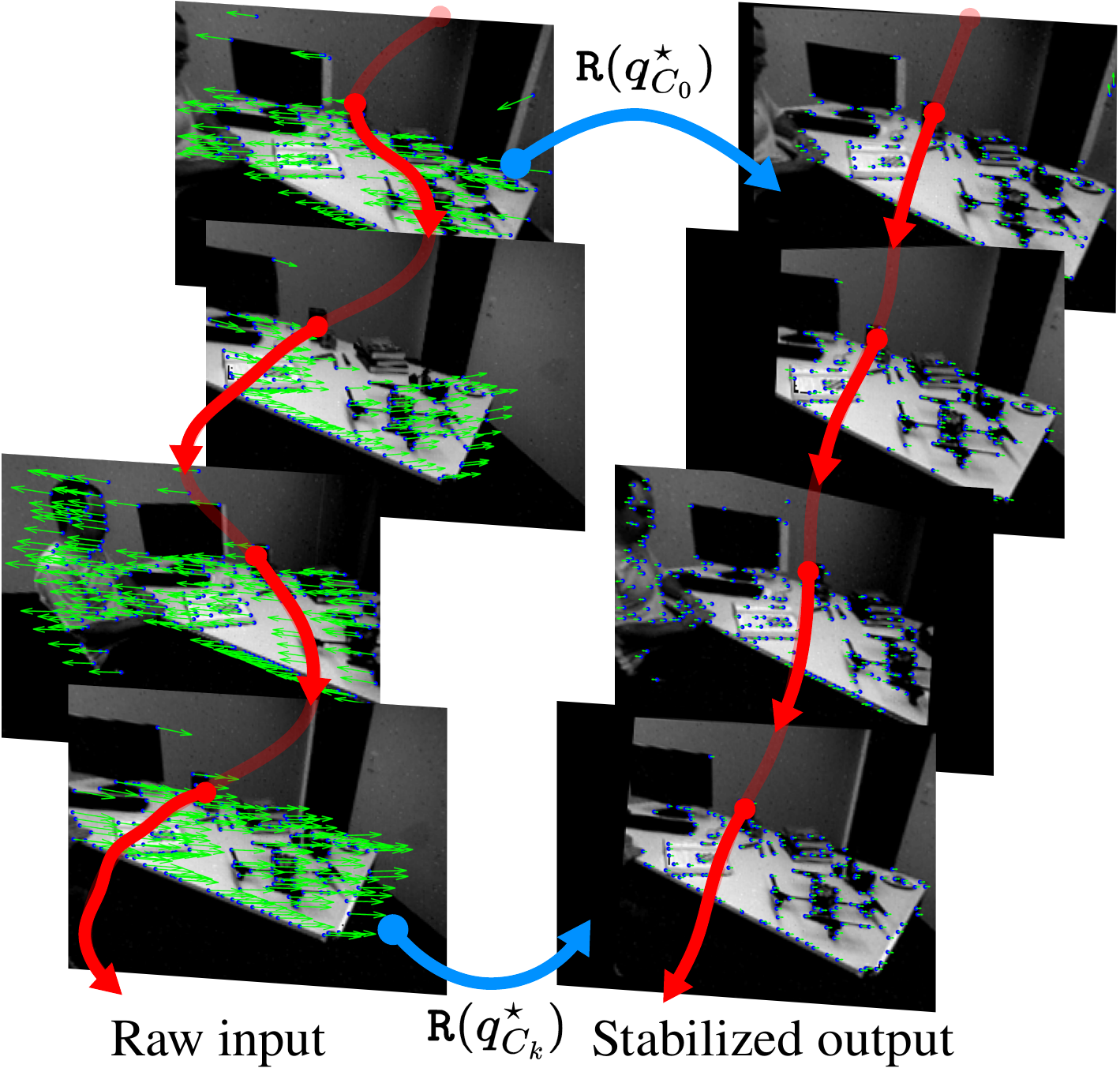}
    \caption{Illustration of visual stabilization applied to images under a dominant rotational motion. 
    The green arrows (optical flow) indicate the magnitude of the motion, which is largely compensated by our orientation stabilization approach. $\Rot(q_{C_{k}}^{\star})$ is the necessary rotation to stabilize frame $k$ (\cref{sec:method:frames}).  
    }
    \vspace{-3ex}
    \label{fig:intro}
\end{figure}

In this paper, we introduce a stabilization approach for frame- and event-based cameras (\cref{fig:intro}). 
It assumes that the camera orientation is available, e.g., from an additional sensor such as an inertial measurement unit (IMU, which is currently integrated into many cameras). 
Our stabilization method is further integrated into frame- and event-based pipelines that estimate the camera linear velocity under the assumption of compensated rotational motion. 
Despite previous stabilization approaches have been presented in the literature, to the best of our knowledge none of them aim at establishing  
the advantages of stabilization for robot perception applications with events or frames. 
Our work is the first time that stabilization on events is demonstrated for high-level perception tasks. 
In the past it was just used for visualization of the event data \cite{Delbruck14iscasshort,Rodriguez22icra}.
A goal of the paper is to show that attitude fusion (e.g., event cameras are often paired with an IMU on the same chip) can help stabilize the visual content and simplify perception pipelines, such as relaxing conditions on subsequent processing methods.

Our contributions can be summarized as follows:
\begin{itemize}
    \item A general framework for visual perception (using events or frames) aided by attitude data fusion to decouple rotational and translational motion estimation (\cref{sec:method}). 
    \item The exemplification of the framework on two tasks: ego-motion (camera velocity) estimation and focal-plane feature tracking (\cref{sec:method}).
    \item Demonstration of the framework on standard datasets \cite{Mueggler17ijrr,Zhu18ral,Gao22ral} 
    and on synthetic data created by a well-known event camera simulator \cite{Rebecq18corl} (\cref{sec:experim}).
    \item An analysis of how stabilization simplifies perception by reducing complexity and processing time, e.g., when estimating the camera's linear velocity (\cref{sec:experim}).
\end{itemize}

The remainder of this letter is organized as follows. 
\Cref{sec:related} summarizes the main works in the topics of the paper. 
\Cref{sec:method} presents the stabilization approach and its integration into a linear velocity estimation pipeline. 
\Cref{sec:experim} provides the experimental results of incorporating stabilization in two perception tasks, and compares the performance against that obtained with non-stabilized data. 
\Cref{sec:conclusion} discusses the trade-off of using stabilization and concludes the paper.

\ifclearlook \cleardoublepage \fi 
\section{Related Work}
\label{sec:related}

\subsubsection*{\textbf{Image stabilization}}
Early image stabilization approaches could be categorized according to how the camera motion was compensated: \emph{mechanically} or \emph{electronically}. 
Mechanical approaches use external hardware to compensate for the changes in the camera's orientation \cite{Oshima89tce}, while electronic techniques typically stabilize image fluctuations through the use of embedded computers and digital zooming \cite{Joon92tce}. 

In the last decades, \emph{Digital Image Stabilization} (DIS) has emerged to stabilize frames without requiring additional hardware. 
Several motion estimation and compensation techniques have been proposed for DIS \cite{Sarigul23survey}. 
The work in \cite{Liu14cvpr} relies on optical flow to estimate the camera motion and compensate for image fluctuations. 
Feature-based methods estimate and compensate the camera motion by extracting and tracking features from frames \cite{Liu2016eccv}, using Kalman filters \cite{Song12tce}, particle filters \cite{Shen09tce}, or path optimization \cite{Grundmann11cvpr}. 
Learning-based approaches utilize deep architectures \cite{Wang19tip}, and learning methods \cite{Xu22tip} 
for the task of video stabilization.

\vspace{1ex}
\subsubsection*{\textbf{Applications}}
The benefits of stabilizing visual data for different applications have been investigated in \cite{Irani94cvpr,Marcenaro01icip,Ling19tcsvt,Yu21cvpr,Shi19icip}. 
The work in \cite{Irani94cvpr} performed frame stabilization to compute camera ego-motion. 
The 2D parametric motion of a static region in the scene was estimated to stabilize the camera's orientation. 
Afterwards, the camera's translation was computed from the focus of expansion of the stabilized frames.
Two image stabilization algorithms for video-surveillance were presented in \cite{Marcenaro01icip}. 
They filtered unwanted motion while preserving the motion of scene objects. 
The experimental results indicated that image stabilization enhanced detection performance compared to non-stabilized surveillance approaches. 
The authors in \cite{Ling19tcsvt} stabilized traffic videos captured by cameras mounted on moving vehicles. 
The approach modeled feature trajectories as a summation of camera motion and object motion, and removed high-frequency motion by solving an optimization problem. 
The work in \cite{Yu21cvpr} presented a learning-based selfie stabilization approach, which stabilized the background and foreground simultaneously using a convolutional network while running at 26 Hz on a GPU. 
Similarly, the work in \cite{Shi19icip} presented a face video stabilization algorithm that removed hand-shake motion on selfie videos, while running on a modern smartphone.  

\begin{table}[t]
\centering
\caption{State-of-the-art frame- and event-based stabilization approaches, 
i.e., the visual input can be events (E) or frames (F).
Frame-based methods correspond to Digital Image Stabilization.
The 2D/3D notation is taken from the survey \cite{Guilluy21spic}.
\label{tab:sota}}
\begin{adjustbox}{max width=\linewidth}
\setlength{\tabcolsep}{4pt}
\begin{tabular}{lccccccc}
\toprule
 & \makecell{Visual\\input} & \makecell{Motion\\estimation} & \makecell{Motion\\model} &  \makecell{CPU\\only} & \makecell{Validated on\\high-level tasks} \\
\midrule
Liu et al.~\cite{Liu14cvpr}    & F & Pixel-based & 2D & \cmark & \xmark \\ 
Liu et al.~\cite{Liu2016eccv}  & F & Feature-based & 2D & \cmark & \xmark\\ 
Wang et al.~\cite{Wang19tip}    & F & Learning-based & 2D & \xmark & \xmark \\
Xu et al.~\cite{Xu22tip}      & F & Learning-based & 2D & \xmark & \xmark \\ 
Delbruck et al.~\cite{Delbruck14iscasshort} & E & External (IMU) & 3D & \cmark & \xmark \\ 
Ours    & E or F & External (IMU) & 3D & \cmark & \cmark \\ 
\bottomrule
\end{tabular}
\end{adjustbox}
\ifclarification
\vspace{-2ex}
\else
\vspace{-3ex}
\fi
\end{table}

In \emph{aerial robotics}, image stabilization is typically used to remove vibrations caused by the platform motion and external disturbances.  
In \cite{merino2012unmanned} an image stabilization approach was proposed to compensate for drone vibrations. 
It performed feature matching and fitting to estimate the camera motion, which was used to eliminate the background displacement between consecutive images.
A video stabilization method for UAVs is proposed in \cite{Odelga17reduas}; it relies on an onboard IMU for estimating the camera orientation. 
Frame stabilization is achieved by applying the homography matrix between the camera and the reference camera orientation. 
The work in \cite{Aswini2021icinpro} presented an image stabilization algorithm for drone surveillance; 
it estimated camera motion from optical flow and compensated video by estimating a smooth version of the camera trajectory. 
The work in \cite{pan2020sj} integrated a mechanical stabilization system for a flapping-wing robot; 
it was able to compensate for the pitch and roll fluctuations caused by the flapping motion of the robot.

\vspace{1ex}
\subsubsection*{\textbf{Event cameras}}
Stabilization with \emph{event cameras} is largely unexplored \cite{Gallego20pami}. 
In \cite{Delbruck14iscasshort} an IMU was hardware-paired with an event camera; 
the angular velocity from the gyroscope was integrated to estimate the camera's orientation, which was used for rotational visual stabilization in the form of motion-compensated events.
Later, the idea of motion-compensating events by maximizing their alignment was leveraged for motion estimation \cite{Gallego17ralshort,Gallego18cvpr}, which could be used for segmentation \cite{Stoffregen19cvpr}, and feature tracking \cite{Rosinol18ral}.
For flapping-wing robots, a stabilization approach using an IMU and motion-compensated events has been initially investigated in \cite{Rodriguez22icra} to compensate for changes in orientation caused by the robot's flapping motion.

\def\figWidth{\linewidth}
\def\customWidthL{0.43\linewidth}
\def\customWidthR{0.53\linewidth}
\begin{figure*}[t]
\centering
\begin{subfigure}{\customWidthR}
  \centering
  \includegraphics[trim={18cm 0.6cm 0 0},clip,width=0.99\linewidth]{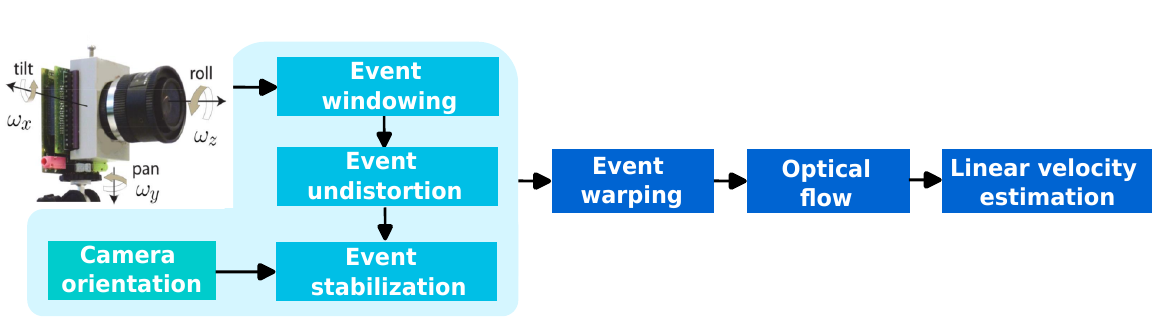}
    \caption{Event-based stabilization.\label{fig:block_diagram:events}}
\end{subfigure}\;\;
\begin{subfigure}{\customWidthL}
  \centering
  \includegraphics[trim={0 1.4cm 20.5cm 0},clip,width=1.05\linewidth]{figures/general_block_diagram_shrink3.pdf}
  \caption{Frame-based stabilization.\label{fig:block_diagram:frames}}
\end{subfigure} 
    \caption{Block diagrams of the proposed stabilization approach, applied to estimation of the camera's linear velocity.}
    \vspace{-3ex}
    \label{fig:block_diagram}
\end{figure*}

\Cref{tab:sota} compares our approach to the most relevant existing works 
highlighting their differences in input data (i.e., frames or events), motion estimation approach, motion model, hardware requirements, and the validation on high-level perception tasks. 
While most previous works aim at solving the problem of stabilizing visual information, very few of them analyze the advantages of stabilization. 
This work aims at unveiling these advantages for the tasks of camera ego-motion (i.e., linear velocity) estimation and feature tracking. 
Our approach effectively stabilizes the changes in camera orientation, enhancing accuracy while reducing processing computational cost.

\ifclearlook \cleardoublepage \fi

\section{Methodology}
\label{sec:method}
Most digital stabilization approaches process consecutive frames to estimate the camera motion 
\cite{Sarigul23survey}.
Despite these works reporting remarkable results by removing fluctuations caused by the camera motion, their estimation techniques are usually computationally expensive, which limits their real-time capabilities and integration on resource-constrained platforms. 
Conversely, other works outsource this task to external sensors (e.g., an IMU \cite{Odelga17reduas}); 
they report impressive results without processing visual information to estimate the motion of the camera. 
We follow the latter philosophy and propose a stabilization approach that assumes the camera orientation is available.  
Hence, we focus on estimating the translational component of the ego-motion, thus assessing the benefits that stabilizing visual data 
has for simpler perception. 

\def\customWidth{0.24\linewidth}
\begin{figure*}[t!]
\centering
\begin{subfigure}{\customWidth}
  \centering
    \gframe{\includegraphics[width=\linewidth]{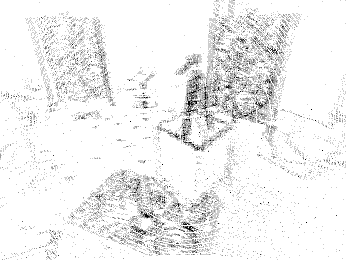}}
    \caption{Undistorted events (shown as an image, without event polarity).\label{fig:datainpipeline:events}}
\end{subfigure}\;
\begin{subfigure}{\customWidth}
  \centering
    \gframe{\includegraphics[width=\linewidth]{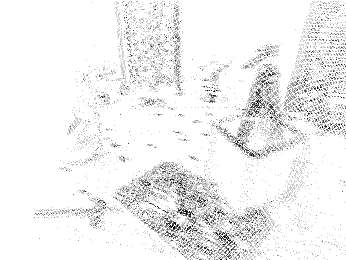}}
    \caption{Events after camera's attitude stabilization.\label{fig:datainpipeline:events:iwestab}}
\end{subfigure}\;
\begin{subfigure}{\customWidth}
  \centering
    \gframe{\includegraphics[width=\linewidth]{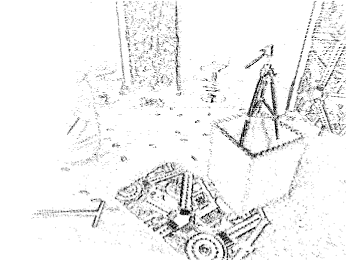}}
    \caption{Events after stabilization and warping (sharp IWE).\label{fig:datainpipeline:events:iwe}}
\end{subfigure}\;
\begin{subfigure}{\customWidth}
  \centering
    \gframe{\includegraphics[width=\linewidth]{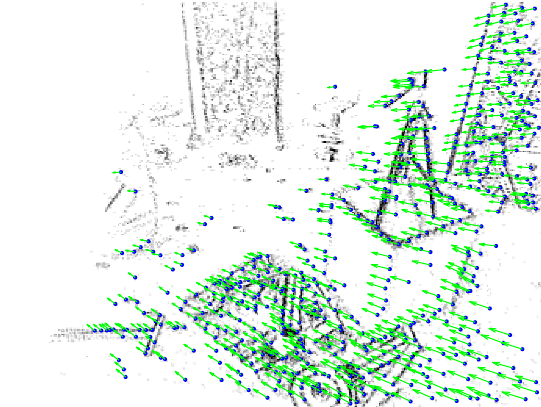}}
    \caption{Optical flow samples passed to ERL-V for velocity estimation.\label{fig:datainpipeline:events:flow}}
\end{subfigure}
\caption{Data in the stabilization processing pipeline of \cref{fig:block_diagram:events} (\cref{subsec:event_warpping}), illustrated using a sequence from \cite{Zhu18ral}.
\label{fig:datainpipeline}}
\vspace{-3ex}
\end{figure*}

\subsubsection*{\textbf{Overview}}
\Cref{fig:block_diagram} shows the block diagrams of our approach to stabilize events and frames.
The event-based stabilization method (\cref{fig:block_diagram:events}) receives as input raw events. 
The pre-processing step first selects events (``windowing'' or ``slicing'') using the area event count criterion from \cite{Liu18bmvc}. 
Next, each event is undistorted and stabilized using a reference camera orientation (that of the last event). 
The event warping module is key to obtain sharp image-like representations of the events (see, \cref{sec:experim}). 
Finally, optical flow is extracted from these event representations and used to estimate the camera's linear velocity by means of ERL-V, which is a simplified version of ERL (Expected Residual Likelihood) \cite{Jaegle16icra}. 
The frame-based method (\cref{fig:block_diagram:frames}) has a similar modular structure: it undistorts and stabilizes consecutive frames to compute optical flow, which is used to estimate the camera's linear velocity using ERL-V.

\vspace{-2.5ex}
\subsection{Event-based stabilization}
\label{sec:method:events}
\Cref{fig:datainpipeline} depicts how events, in the form of Images of Warped Events (IWE), evolve as they are processed by the stabilization pipeline of \cref{fig:block_diagram:events}. 

\subsubsection{\textbf{Camera motion estimation}}
Our approach assumes that camera orientation is available. 
This is practical because many modern event cameras come with an integrated IMU (hardware synchronized).
For instance, camera orientation can be obtained by integrating the gyroscope measurements of the IMU. 
However, this approach is prone to suffer from drift. 
A more robust approach consists in fusing gyroscope and accelerometer data to estimate the camera orientation. 
The quaternion-based method in \cite{Valenti2015sensors} 
reports remarkable results on estimating the IMU's orientation by fusing also
magnetometer measurements in a complementary filter 
(the magnetometer confers robustness by correcting the heading drift on the $Z$ (i.e., gravity) axis of the IMU). 
The IMU orientation is provided in quaternion form, $q_{I} = [q_w, q_x, q_y, q_z]$, 
and the camera orientation is given by $q_{C}=q_{o} q_{I} q_{o}'$, where $q_{o}$ represents the rotation between the camera frame and the IMU, and $q_{o}'$ stands for the quaternion conjugate of $q_{o}$.

\vspace{1ex}
\subsubsection{\textbf{Event pre-processing}} 
Each event $e_k$ is described by a tuple $(\bx_k, t_k, p_k)$, where $\bx_k = (x_k,y_k)^\top$ corresponds to the pixel coordinates, $t_k$ is the timestamp (with \si{\micro\second} resolution), and $p_k\in \{+1,-1\}$ is the polarity (sign) of the brightness change. 

\emph{Windowing and Undistortion}.
Different from classical approaches that select events either by a fixed count of them or by a fixed time interval duration, we adopt the method in \cite{Liu18bmvc}. 
It subdivides the image plane into a tile of regions, and selects events until the count of them in any region exceeds a preset threshold. 
This approach provides asynchronous windows/packets of events while adapting to the scene dynamics and texture by measuring the event density in different regions of the scene. 
Afterwards, events are efficiently undistorted using a look-up table built from the known camera calibration.
Undistorted events are displayed in \cref{fig:datainpipeline:events}.

\emph{Stabilization}. 
We compensate for fluctuations in camera orientation $q_{C}$ with respect to a reference rotation $q_{C_{0}}$. 
Although $q_{C_{0}}$ is ideally constant, it may vary when the stabilized visual data moves out of the field of view (FOV). 
In this situation, we update $q_{C_{0}}$ to the latest camera orientation, which produces a ``saccadic'' effect on the output while stabilizing data for the full camera trajectory. 
The process of updating $q_{C_{0}}$ is detailed in \cref{sec:experim}.
The quaternion $q_{C}^{\star} = (q_{C_{0}})^{-1}q_{C}$ defines the necessary rotation to move the current window of events to the reference $q_{C_{0}}$.
Each event is orientation-stabilized (in homogeneous coordinates) by:
\begin{equation}
    \bx_k' \sim \Kint\, \Rot(q_{C}^{\star})\, \Kint^{-1}\bx_k,
    \label{eq:event_stabilization}
\end{equation}
where $\Kint$ is the intrinsic parameter matrix, and $\Rot(q_{C}^{\star})$ is the rotation 
with $q_{C}^{\star} = (q_w, q_x, q_y, q_z)$.
Event stabilization is illustrated in the processing step between \cref{fig:datainpipeline:events} and \cref{fig:datainpipeline:events:iwestab}. 

\vspace{1ex}
\subsubsection{\textbf{Event warping}}
\label{subsec:event_warpping}
The orientation-stabilization in \eqref{eq:event_stabilization} happens at the window level. 
A finer stabilization, called ``contrast maximization'' or ``motion compensation'' 
\cite{Gallego17ralshort} happens within each window. 
Each event is warped according to a different velocity vector so as to produce sharp IWEs \cite{Gallego18cvpr}:
\begin{equation}
\textstyle 
I(\bx) = \sum_{k} \mathcal{N}(\bx; \bx^{\prime\prime}_k, \sigma^2),
\label{eq:IWE}
\end{equation}
where $\mathcal{N}(\bx; \bx^{\prime\prime}_k, \sigma^2)$ is a bivariate Gaussian function of width $\sigma=1$ px centered at the warped event $\bx^{\prime\prime}_k = \mathbf{W}(\bx'_k; t_k, \boldsymbol{\theta})$.
Event warping (after stabilization) is illustrated in \cref{fig:datainpipeline:events:iwe}.

We use the state-of-the-art method in \cite{Shiba22eccv} to compute the sharp IWEs, 
which are then fed to the last two blocks of the pipeline in \cref{fig:block_diagram:events}: optical flow computation on (stabilized) IWEs and camera velocity estimation.
The green arrows in the last IWE plot of \cref{fig:datainpipeline:events:flow} denote optical flow vectors.

We also consider \emph{time surfaces} (TS) \cite{Lagorce17pami},
which are a simple and effective representation, storing the time of the last event at each pixel. 
TSs with exponential kernels (of time constant $\tau$) highlight the occurrence of recent events. 
To provide a TS representation that adapts to the duration of the event windows, we propose an approach that adjusts the value of $\tau$ based on the time span $\Delta t$ of each event window. 
A low-pass filter updates the time decay reference as $\tau_j = (1-\alpha)\Delta t + \alpha \tau_{j-1}$ (with $\alpha=0.3$ in the experiments). 

\vspace{1ex}
\subsubsection{\textbf{Linear velocity estimation}} 
Our approach estimates the camera ego-motion from optical flow. 
It is inspired by \cite{Jaegle16icra}, which solves the problem of camera ego-motion estimation in the presence of noisy, sparse optical flow. 
We selected ERL for its remarkable performance and adaptability to perform with frames, TSs, and IWEs, which serves as a suitable pipeline for our comparison differently to solely event-based linear velocity estimators \cite{Xu2024TRO,Lu2024rss}.
Let us describe ERL \cite{Jaegle16icra} and present our modification to estimate camera ego-motion under the assumption that the angular velocity is known and has been compensated for (thus it is equivalent to a purely translating camera, i.e., $\angvel=0$).

ERL aims at computing the weight confidence for each input optical flow sample, and uses a lifted robust kernel to jointly estimate the weight and the motion parameters. 
Assuming zero angular velocity ($\angvel=0$), the apparent velocity of a static scene point is due to the camera's linear velocity, the scene depth, and the projected point's location:
\begin{equation}
\flowvec(\bx_{i}) = \systemMatrix(\bx_{i})\linvel \invdepth_{i}(\bx_{i}),
\label{eq:motion_model}
\end{equation}
where $\flowvec_i \equiv \flowvec(\bx_i) = (u_i,v_i)^{\top}$ corresponds to the optical flow at the projected point $\bx_i =(x_i,y_i)^{\top}$, $\linvel = (V_x,V_y,V_z)^{\top}$ is the camera linear velocity, 
and $\invdepth_{i}\equiv \invdepth(\bx_i) > 0$ is the inverse of the scene depth at $\bx_i$. 
Matrix $\systemMatrix(\bx_{i})$ is $2\times 3$, corresponding to the translational part of the feature sensitivity matrix \cite{Corke17book}.

The ego-motion estimation problem consists in computing $\linvel$ and the inverse depth values $\{\invdepth_i\}_{i=1}^{N}$ from the input optical flow velocities $\{\flowvec_i\}_{i=1}^{N}$ at locations $\{\bx_i\}_{i=1}^{N}$. 
Since the unknowns $\linvel$ and $\invdepth_i$ are multiplying each other in \eqref{eq:motion_model}, the camera linear velocity can only be recovered up to a scale factor (i.e., the well-known scale factor ambiguity in monocular SLAM, represented in the scene by the inverse depths $\invdepth_i$). 
Using vector notation $\invdepthvec = (\invdepth_1,\ldots,\invdepth_N)^\top$, 
$\flowvec = (\flowvec^\top_{1},\ldots,\flowvec^\top_{N})^\top$ and algebraic operations, 
the problem of minimizing the square residuals of \eqref{eq:motion_model} can be rewritten as:
\begin{equation}
\min_{\linvel,\invdepthvec} \| \tilde{\systemMatrix}(\linvel)\invdepthvec - \flowvec \|^{2}_{2}, \quad\text{s.t.}\quad \| \linvel \|=1.
\end{equation}
where $\tilde{\systemMatrix}(\linvel) \doteq \operatorname{diag}(\systemMatrix(\bx_1)\linvel,\ldots,\systemMatrix(\bx_N)\linvel)$ is a $2N\times N$ matrix. Following \cite{Jaegle16icra}, we first solve for $\invdepthvec$ using least-squares:
\begin{equation}
\min_{\linvel, \| \linvel \|=1} \min_{\invdepthvec} \| \tilde{\systemMatrix}(\linvel)\invdepthvec - \flowvec\|^{2}_{2} \\
= \min_{\linvel, \| \linvel \|=1} \| \tilde{\systemMatrix}(\linvel)^{\bot} \flowvec\|^{2}_{2},
\label{eq:cost_function}
\end{equation}
where $\tilde{\systemMatrix}(\linvel)^{\bot}$ is the orthogonal complement of $\tilde{\systemMatrix}(\linvel)$.  
This expression is fast to compute as $\tilde{\systemMatrix}(\linvel)$ is sparse and does not depend on $\invdepthvec$. 
Hence, one may estimate $\linvel$ by minimizing \eqref{eq:cost_function}. 
However, this approach is not stable in the presence of outliers. 
The work in \cite{Jaegle16icra} proposes a robust approach introducing a confidence weight $w({\bx_i}) \in [0,1]$ for each flow vector:
\begin{equation}
\min_{\linvel,\| \linvel \|=1} \| w \circ \tilde{\systemMatrix}(\linvel)^{\bot} \flowvec \|^{2}_{2}. 
\end{equation}
Our modification of ERL, called ERL-V, consists in simplifying the formulation for the case $\angvel=0$ (i.e., avoid unnecessary computations in \eqref{eq:motion_model}) and feed the corresponding optical flow.

\vspace{-2.5ex}
\subsection{Frame-based stabilization}
\label{sec:method:frames}

The block diagram for frame-based stabilization (\cref{fig:block_diagram:frames}) has similarities with that in \cref{fig:block_diagram:events}.
Frames are stabilized for the changes in the camera orientation $q_{C}$ after removing radial and/or tangential distortion (due to the lens).
The homography $\mH \sim \Kint\,\Rot(q_{C}^{\star})\,\Kint^{-1}$ 
is applied to the input frames for stabilization.
Note that there is no ``within window stabilization'' like in the event-based case because all pixels of the frame have the same timestamp.
Sparse optical flow is then computed (e.g., using \cite{Lucas81ijcai}) from stabilized frames and used for linear camera velocity estimation (ERL-V).

\ifclearlook \cleardoublepage \fi
\section{Experiments}
\label{sec:experim}

Let us assess the benefits of visual data stabilization on two tasks. 
First, we evaluate the performance of the proposed linear velocity estimation approach using images and events (\cref{sec:linearvelestination}). 
The evaluation is carried out on synthetic and standard event-based datasets \cite{Mueggler17ijrr,Zhu18ral}. 
Second, we evaluate the benefits in the context of feature tracking (\cref{sec:experim:featuretracking}),
comparing the performance of frame- and event-based methods on stabilized and non-stabilized data.

\vspace{-2.5ex}
\subsection{Linear velocity evaluation}
\label{sec:linearvelestination} 
First, we test the problem of estimating the camera's linear velocity. 
The hyperparameter configuration of the stabilization pipeline used in the experiments 
is empirically determined based on the camera motion and the event dynamics. 
In the event \emph{windowing} \cite{Liu18bmvc}, we divide the image plane into a tile of $34 \times 26$ regions and set the counting threshold to $50$ events (for dataset \cite{Zhu18ral}), $200$ (for \cite{Mueggler17ijrr}), and $500$ (for \cite{Gao22ral}) due to the event occurrence in each dataset. 
This produces a good distribution of events on the image plane for event warping. 
The reference rotation $q_{C_{0}}$ is updated when the displacement of the image centroid surpasses $W/6$ \SI{}{px} 
by applying $q_{C}^{\star}$ to frames or events, where $W$ is the sensor width. 
This configuration keeps a large part of the image plane in the FOV of the stabilized camera while avoiding unnecessary saccades on the output of the pipeline. 
For the TS, the time decay constant $\tau_0$ is set to \SI{35}{\milli\second} for data in \cite{Mueggler17ijrr}, and \SI{100}{\milli\second} for \cite{Zhu18ral} and \cite{Gao22ral}. 
Similarly to the IWEs, $\tau_0$ is selected based on the event occurrence on each dataset. 
Finally, we used the well-known algorithm \cite{Lucas81ijcai} to extract optical flow from consecutive image representations. 
\cref{fig:visualization_stabilization} shows a sample frame and an IWE after stabilization.

\def\figWidth{0.5\linewidth}
\def\customWidthL{0.4\linewidth}
\def\customWidthR{0.4\linewidth}
\begin{figure}[b!]
\centering
\vspace{-2ex}
\begin{subfigure}{\customWidthR}
  \centering
  \includegraphics[trim={0cm 1.5cm 0 2.0cm},clip,width=0.99\linewidth]{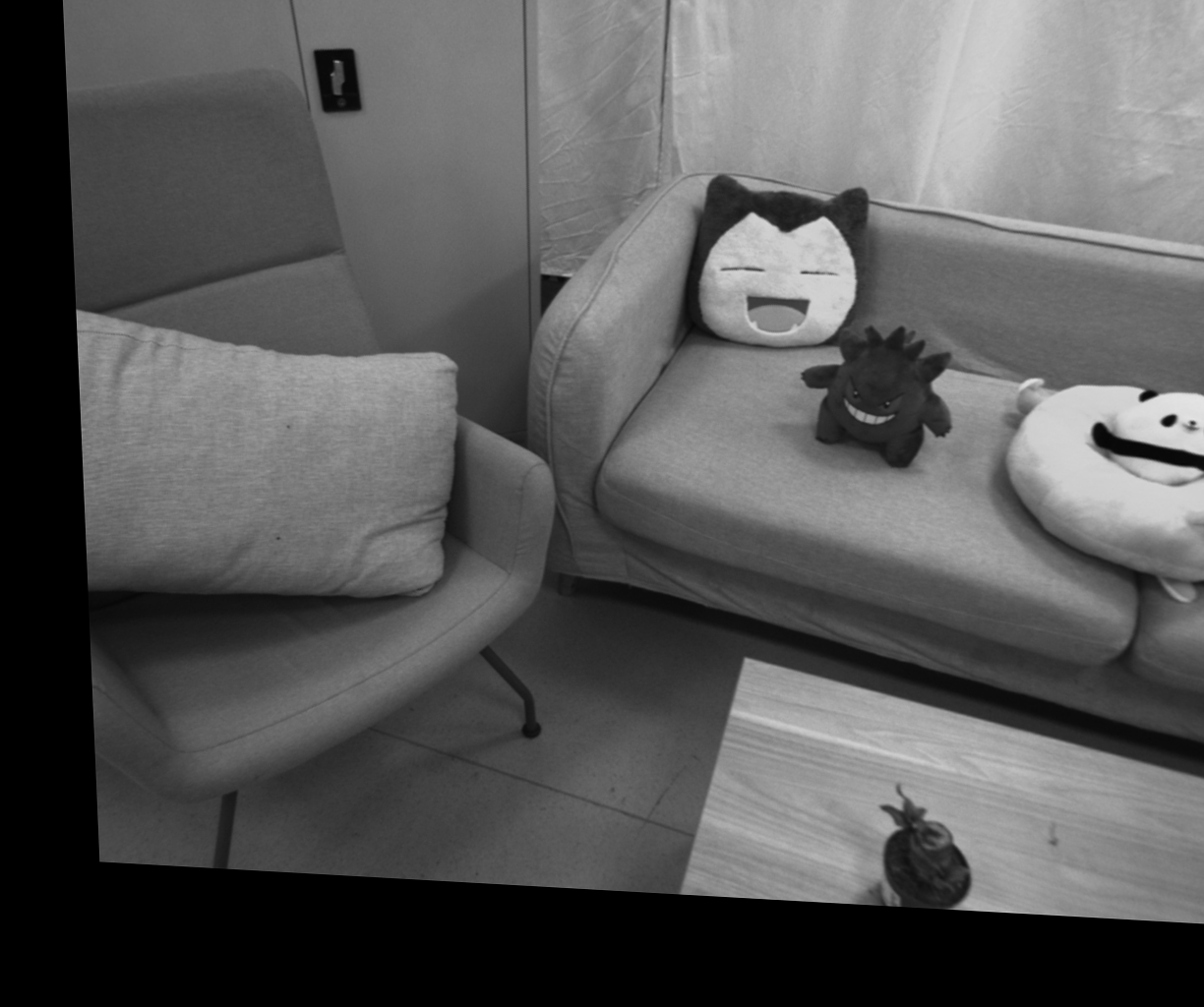}
\end{subfigure}\;\;
\begin{subfigure}{\customWidthL}
  \centering
  \gframe{\includegraphics[trim={0 0 0cm 0},clip,width=0.99\linewidth]{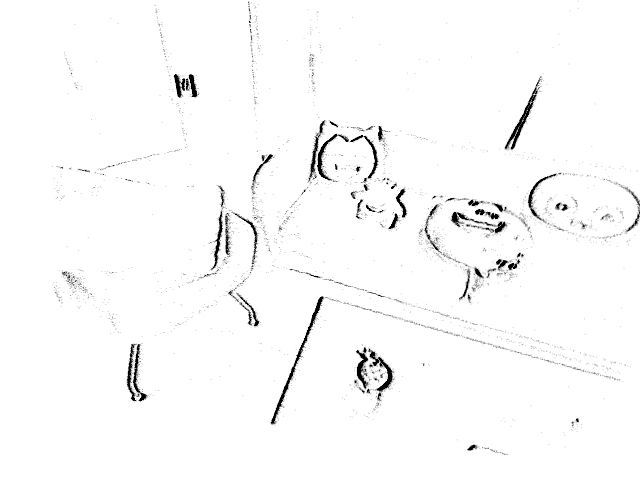}}
\end{subfigure} 
    \caption{Visualization of a frame (left) and an IWE (right) after stabilization. 38
    Data from VECtor3 sequences \cite{Gao22ral}.}
    \label{fig:visualization_stabilization}
\end{figure}

\subsubsection{\textbf{Experiments on synthetic data}}
\label{sec:experim:syntheticdatasets}

\begin{figure}[b]
    \vspace{-2ex}
    \begin{center}
    \includegraphics[trim={2.5cm 0.7cm 1.0cm 0.7cm},clip,width=1.0\linewidth]{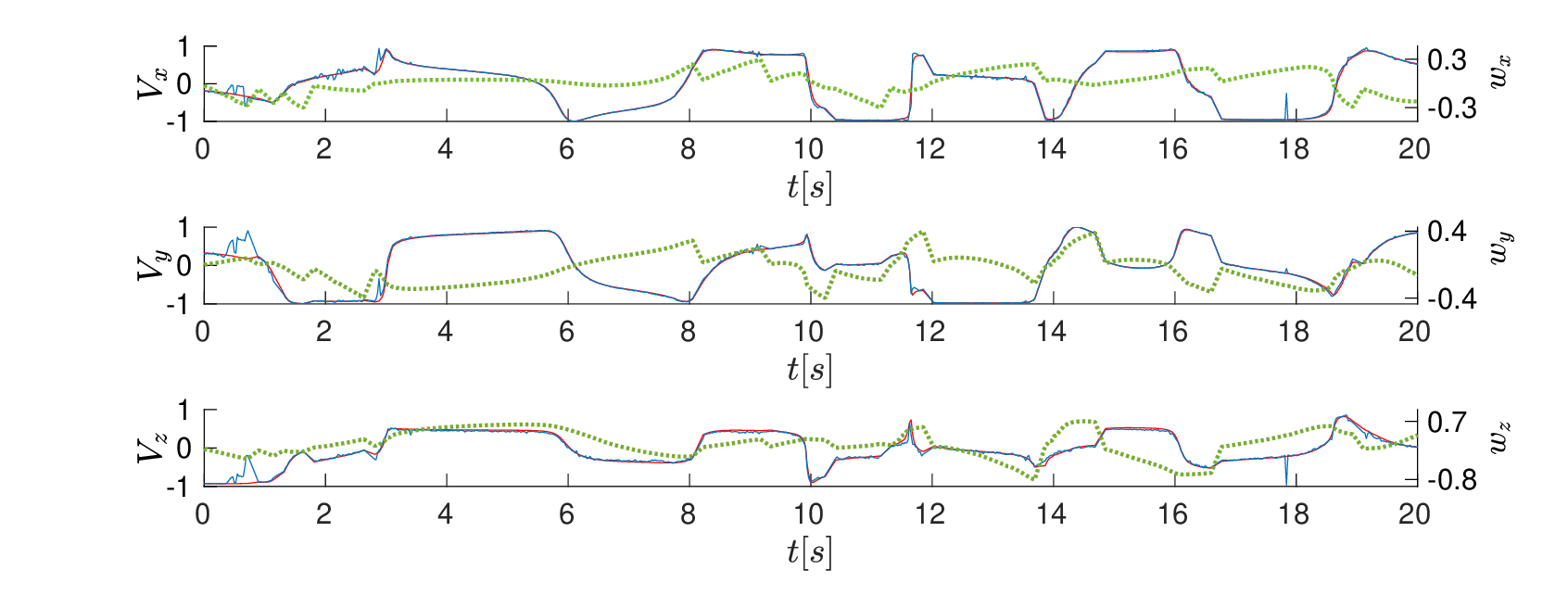}
    \caption{Velocity plots using ERL-V with IWEs on sequence \emph{ESIM 3}.
    Overall, estimation is good, producing small errors. 
    The red curves correspond to the ground truth velocities, while the blue curves describe the estimated $\linvel$. 
    The green dotted curves represent the camera angular velocities over time.}
    \vspace{-1ex}
    \label{fig:validation_esim}
    \end{center}
\end{figure}

A synthetic stabilization dataset is collected using the Event Camera Simulator (ESIM) \cite{Rebecq18corl} using the default sensor parameters. 
Three sequences were created describing challenging 6 degrees-of-freedom (DOF) motions where stabilization can play a key role in simplifying robot perception. 
\Cref{tab:experim:ourdata} summarizes the linear velocity estimation results of using ERL (unstabilized) and our method ERL-V with grayscale frames (\cref{fig:block_diagram:frames}), and event-based IWEs, and TSs (\cref{fig:block_diagram:events}). 
Besides, the GT column corresponds to the results obtained by using the ESIM's ground truth optical flow with ERL.
We use as evaluation metric the Mean Absolute Error (MAE) angle between the estimated linear velocity $\linvel$ and the ground truth (GT) velocity $\linvel_\text{GT}$ 
(normalized to unit norm, since scale is unobservable). 
The results evidence the benefits of stabilization: 
the proposed approach using IWEs provides the lowest MAE even on the most complex sequence \textit{ESIM 3}, 
where fast and sudden changes in the camera motion hinder 
the estimation of $\linvel$ (see DOF plots in \cref{fig:validation_esim}).

\begin{table}[t]
\caption{Linear velocity estimation on synthetic sequences.
Mean Absolute Error (MAE) [$^\circ$].
Best in bold; 2nd underlined.\label{tab:experim:ourdata}}
\vspace{-1ex}
\centering
\begin{adjustbox}{max width=\linewidth}
\begin{tabular}{lrrrrrrr}
\toprule
\multirow{2}{*}{Sequence} & & \multicolumn{3}{c}{ERL (non-stab.)}  & \multicolumn{3}{c}{ERL-V (stab.)} \\ 
\cmidrule(l{1mm}r{1mm}){3-5}
\cmidrule(l{1mm}r{1mm}){6-8}
& GT & {Frames} &  {TS} & IWE &  {Frames} & {TS} &  IWE \\
\midrule
\emph{ESIM 1} & 1.45 & \underline{1.81} & 3.15 & 2.69 & 1.91 & 3.76 & \textbf{1.46} \\ 
\emph{ESIM 2} & 1.46 & \underline{2.36} & {8.27} & 2.69 & {3.67} & {6.05} & \textbf{1.88} \\ 
\emph{ESIM 3} & 1.52 & \underline{3.23} & {13.65} & 3.73 & {4.41} & {10.90} & \textbf{3.22} \\ 
\bottomrule             
\end{tabular}
\end{adjustbox}
\vspace{-3ex}
\end{table}

\subsubsection{\textbf{Results on standard datasets}}
\label{sec:experim:standarddatasets}

The ECD dataset \cite{Mueggler17ijrr} 
provides 6-DOF perception measurements collected with a DAVIS240C camera ($240\times 180$ px resolution). 
It describes sequences collected under different motion, illumination, and scene conditions.
The MVSEC dataset \cite{Zhu18ral} consists of data sequences from several perception sensors, including a pair of stereo DAVIS346 ($346\times 260$ px). 
It was recorded on board different moving vehicles: a drone, a car, and a motorcycle. 
The VECtor dataset \cite{Gao22ral} provides synchronized measurements from several sensors, including stereo event- and standard cameras, and a high-performance IMU. 
All datasets are used as standard benchmarks in several tasks, such as event-based feature tracking, optical flow, pose estimation, and SLAM \cite{Gallego20pami}.

In most of these experiments, the camera orientation is provided by the GT poses from a motion capture system. 
We preferred this approach over estimating the camera orientation from the IMU of ECD and MVSEC for two main reasons:
(i) the strong variations in the camera orientation in \cite{Mueggler17ijrr} tend to produce large heading drifts using only the available accelerometer and gyroscope measurements, and
(ii) the IMU data in \cite{Zhu18ral} are considerably noisy (since it was a prototype event camera that heated up considerably), which hinders the estimation of the camera orientation.  
Nevertheless, experiments with VECtor sequences include stabilization with both GT and IMU orientations. The goal of these experiments 
is to analyze the advantages of stabilization assuming the input orientation is sufficiently accurate (\cref{sec:method}), 
thus limiting the propagation of errors in camera orientation.

\Cref{tab:experim:publicdata} compares camera ego-motion estimation performance using raw data with ERL vs.~using our approach (ERL-V). 
The evaluation considers the \emph{boxes}, \emph{dynamic}, and \emph{poster} \emph{6DOF} sequences in \cite{Mueggler17ijrr} (1-minute long), the \emph{indoor\_flying1-3} sequences in \cite{Zhu18ral} ($>$ 90 s long), and \emph{sofa\_normal}, \emph{desk\_normal}, and \emph{mountain\_normal} sequences in \cite{Gao22ral} ($>$ 60 s long). 
All sequences describe 6-DOF camera motions, which allow stabilizing for the changes in the camera orientation and estimating the camera's translational motion.  
The results in \cref{tab:experim:publicdata} demonstrate that in most cases stabilizing visual data provides better ego-motion estimations. 

\textbf{ECD}.
Among the sequences in \cite{Mueggler17ijrr}, \emph{poster 6DOF} presents the most challenging scenario for stabilization. 
The textures in the scene lie on a plane, which is a challenging situation for ego-motion estimation pipelines based on optical flow \cite{Irani94cvpr}. 
Nevertheless, our approach ERL-V provides valid estimations using both frames or events under these conditions by reducing the complexity of the linear velocity computation problem.

\textbf{MVSEC}.
Similarly, the results on the \emph{indoor\_flying} sequences evidence that stabilizing visual data can produce more accurate linear velocity estimations. 
ERL-V using the TS gives less accurate results because the dataset contains many pixels that do not trigger events (due to the prototype DAVIS sensors used), and when they are converted into a stabilized time surface they produce artifacts that harm the optical flow estimator.
By contrast, ERL-V using IWEs reports the highest accuracy; 
it provides image representations without motion blur (which otherwise affects the grayscale APS frames).
Besides, the performance of our ERL-V method using IWEs is close to the results obtained using the GT optical flow in \cite{Zhu18ral} and ERL. 
This evidences the potential advantages of event-based vision over frame-based methods for stabilization.

\begin{figure}[t]
    \begin{center}
    \includegraphics[trim={2.5cm 0.7cm 0.9cm 0.7cm},clip,width=1.0\linewidth]{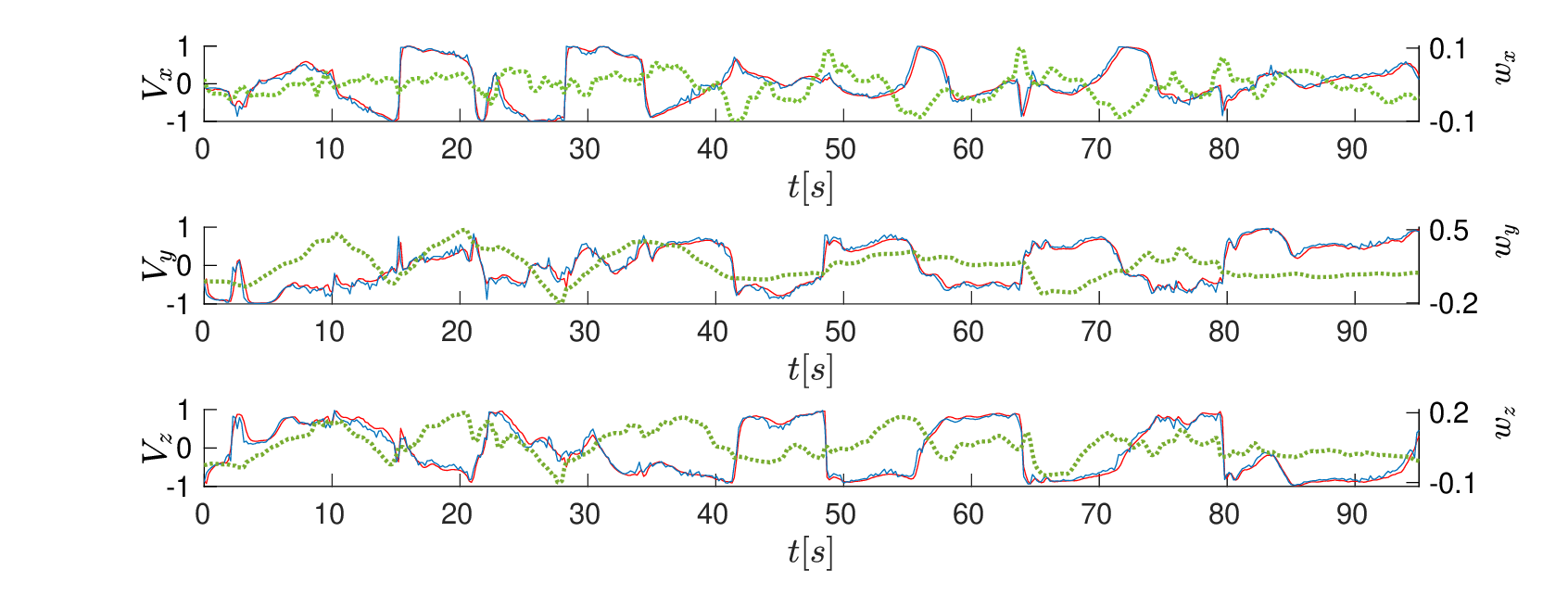}
    \caption{Results on the MVSEC \flythree{} sequence using our event-based stabilization approach. The red and blue curves represent the GT and the estimated linear velocities. 
    The green curves correspond to the angular camera velocities.}
    \vspace{-4ex}
    \label{fig:validation_MVSEC}
    \end{center}
\end{figure}

\begin{table}[b]
\vspace{-3ex}
\caption{Linear velocity estimation on real-world sequences from \cite{Mueggler17ijrr,Zhu18ral,Gao22ral}.
Mean Absolute Error (MAE) [$^\circ$].
\label{tab:experim:publicdata}}
\vspace{-2ex}
\begin{center}
\begin{adjustbox}{max width=\linewidth}
\setlength{\tabcolsep}{4pt}
\renewcommand{\arraystretch}{1.15}
\begin{threeparttable}
\begin{tabular}{lrrrrrrrr}
\toprule
\multirow{2}{*}{Sequence} & & \multicolumn{3}{c}{ERL (non-stab.)} & \multicolumn{3}{c}{ERL-V (stab.)} & \\ 
\cmidrule(l{1mm}r{1mm}){3-5}
\cmidrule(l{1mm}r{1mm}){6-8}
& GT & {Frames} &  {TS} & IWE &  {Frames} & {TS} &  IWE & ESVIO \\ \midrule
\emph{shapes 6DOF} \cite{Mueggler17ijrr}\tnote{*} & -- & {41.24} & {45.56} & 42.50 & \textbf{15.91} & {22.09} & \underline{17.82} & -- \\ 
\emph{poster 6DOF} \cite{Mueggler17ijrr}\tnote{*} & --  & {35.58} & {46.56} & 54.13 & \textbf{11.04} & {21.75} & \underline{11.42} & -- \\ 
\emph{boxes 6DOF} \cite{Mueggler17ijrr}\tnote{*} & -- &{\textbf{6.19}} & {19.94} & 9.98 & \underline{8.35} & {31.04} & 11.83 & -- \\ 
\emph{dynamic 6DOF}\cite{Mueggler17ijrr}\tnote{*} & -- & {18.18} & {24.63} & 17.67 & \textbf{13.26} & {20.37} & \underline{14.24} & -- \\ 
\flyone{} \cite{Zhu18ral} & 4.00 & {10.88} & {16.72} & 13.56 & \underline{10.65} &  {45.67} & \textbf{9.05} & 15.53 \\ 
\flytwo{} \cite{Zhu18ral} & 7.68 & {8.82} &{14.35} & 12.17 & \underline{8.37} & {46.16} & \textbf{7.88} & 12.85 \\ 
\flythree{} \cite{Zhu18ral} & 6.56 & \underline{8.02} & 13.68 & 11.89 & 8.06 & 44.52 & \textbf{7.78} & 15.77 \\ 
{\sofanormal{} \cite{Gao22ral}} & {--} &{\underline{5.08}} & {17.45} & {12.31} & {10.53} & {9.22} & {{5.54}} & {56.72} \\
{\sofanormalimu{}} & {} & {} & {} & {} & {9.03} & {8.03} & {\textbf{4.84}} & \\
{\desknormal{} \cite{Gao22ral}} & {--} & {\underline{10.12}} & {17.03} & {14.03} & {13.63} & {16.98} & {{11.85}} &{46.07} \\
{\desknormalimu{}} &  &  &  &  & {11.59} & {14.95} & {\textbf{8.39}} & \\
{\mountainnormal{} \cite{Gao22ral}} & {--} & {\underline{9.72}} & {16.52} & {13.01} & {14.56} & {15.21} & {{9.81}} & {52.78} \\
{\mountainnormalimu{}} &  &  &  & {} & {14.19} & {14.60} & {\textbf{8.24}} &  \\
\bottomrule
\end{tabular}
\begin{tablenotes}
\item[*] Only the first \SI{30}{\second}  of each sequence were considered for evaluation due to the motion blur limitation in the APS frames, which affects optical flow estimations using \cite{Lucas81ijcai}.
\vspace{-3ex}
\end{tablenotes}
\end{threeparttable}
\end{adjustbox}
\end{center}
\end{table}

\Cref{fig:validation_MVSEC} shows the result of estimating $\linvel$ using our event-based approach on the \emph{indoor\_flying3} sequence. 
Our method returns small errors even under the agile motion performed by the drone. 
The estimation errors shown in the figure are caused by low-textured regions, where having few optical flow measurements cannot properly describe the camera motion. 
Besides, linear velocity estimation error increases when the angular velocity suddenly changes its direction.

\textbf{VECtor}. Different from the previous analysis, these experiments include event and frame stabilization results using attitude estimations from an external IMU. The inertial sensor in \cite{Gao22ral} provides excellent attitude estimations, with roll, pitch, and yaw RMSE of \SI{0.34}{\degree}, \SI{0.56}{\degree}, and \SI{0.71}{\degree}, respectively. In this case, stabilizing using IMU orientation provides more accurate estimations than using the GT. This is directly related to the higher frequency of the IMU (\SI{200}{\hertz}) compared to the GT (\SI{120}{\hertz}), which reduces the temporal difference between the attitude estimations and the visual data. 
Frame stabilization provides lower performance than using ERL similar to \emph{boxes\_6DOF} and \emph{indoor\_flying3}. Conversely, ERL-V reports more accurate linear velocity estimations with IWEs and TS than ERL, while utilizing samples stabilized with the IMU. However, ERL-V performs almost equally to ERL using lower frequency attitude references, which indicates the relevance of the input orientation's temporal resolution for stabilization. 

\subsubsection{\textbf{Comparison with an event-based stereo visual-inertial odometry method}}
We assess ERL's performance against ESVIO \cite{Chen23ral}. 
First, we used ESVIO trajectories for MVSEC and VECtor sequences and aligned them with their GT poses, as described in \cite{Chen23ral}. 
Afterwards, the camera's linear velocity was computed for each aligned pose by following the approach in \cite{Jaegle16icra}. 
It is worth noticing that we do not compare ESVIO with ERL-V, as the VIO method does not work on stabilized events. 
ESVIO's performance is reported in \cref{tab:experim:publicdata}. 
In most cases, ERL outperforms ESVIO using frames, TSs, and IWEs. 
While the event-based VIO method performs close to ERL on MVSEC sequences, it reports significant errors on VECtor sequences due to important differences between ESVIO's estimated camera orientation and the GT.

\begin{table*}[t!]
\caption{Feature tracking evaluation using KLT (frames) and EKLT (events) for the raw data from \cite{Mueggler17ijrr}.}
\vspace{-1ex}
\centering
\begin{adjustbox}{max width=\linewidth}
\setlength{\tabcolsep}{5pt}
\begin{tabular}{lrrrrrrrrrr|rrrrrrrrrr}
\toprule
\multirow{2}{*}{Sequence} & \multicolumn{5}{c}{Frames} & \multicolumn{5}{c}{Stabilized frames} & \multicolumn{5}{c}{Events} & \multicolumn{5}{c}{Stabilized events} \\ 
\cmidrule(l{1mm}r{1mm}){2-6}
\cmidrule(l{1mm}r{1mm}){7-11}
\cmidrule(l{1mm}r{1mm}){7-11}
\cmidrule(l{1mm}r{1mm}){12-16}
\cmidrule(l{1mm}r{1mm}){17-21}
& {TE$\downarrow$} &  {TTE$\downarrow$} & {ETE$\downarrow$} &  {NFA$\uparrow$} & {TFA$\uparrow$} &  {TE$\downarrow$} & {TTE$\downarrow$} & {ETE$\downarrow$} & {NFA$\uparrow$} &  {TFA$\uparrow$} & {TE$\downarrow$} & {TTE$\downarrow$} & {ETE$\downarrow$} & {NFA$\uparrow$} &  {TFA$\uparrow$} &  {TE$\downarrow$} & {TTE$\downarrow$} & {ETE$\downarrow$} & {NFA$\uparrow$} &  {TFA$\uparrow$} \\
\midrule
\emph{shapes 6DOF} & 0.56 &  0.45 & 0.77 & 0.95 & 0.95 & 0.28 & 0.22 & 0.41 & 0.94 & 0.91 & 0.79 &  0.46 & 0.75 & 0.84 & 0.61 & 0.76 & 0.31 & 0.47 & 0.84 & 0.59 \\ 
\emph{shapes rotation} & 0.41 &  0.36 & 0.85 & 0.99 & 0.91 & 0.21 & 0.19 & 0.34 & 0.98 & 0.98 & 0.68 &  0.32 & 0.75 & 0.84 & 0.76 & 0.66 & 0.19 & 0.54 & 0.85 & 0.78 \\ 
\emph{poster 6DOF} & 0.48 &  0.46 & 0.89 & 1.00 & 1.00 & 0.41 & 0.38 & 0.94 & 0.97 & 0.95 & 0.58 &  0.45 & 0.82 & 0.78 & 0.75 & 0.62 & 0.29 & 0.67 & 0.74 & 0.68 \\ 
\emph{poster rotation} & 0.29 &  0.26 & 0.73 & 0.98 & 0.97 & 0.23 & 0.21 & 0.47 & 0.97 & 0.95 & 0.49 &  0.22 & 0.55 & 0.79 & 0.64 & 0.52 & 0.13 & 0.39 & 0.63 & 0.57 \\ 
\bottomrule             
\end{tabular}
\end{adjustbox}
\vspace{-3ex}
\label{tab:experim:tracking-all}
\end{table*}

\subsubsection{\textbf{Processing time reduction}}
\label{sec:experim:time_reduction}
The process of undistorting and stabilizing an image and an event takes around \SI{10}{\milli\second} (1224 $\times$ \SI{1024}{px}) and \SI{0.12}{\micro\second}, respectively. 
That is, our approach runs faster than the typical frame rate (i.e., \SI{30}{\hertz}) of standard cameras, and processes up to \SI{8.33}{\si\mega eps} satisfying the event throughput in some fast robotic applications \cite{RodriguezGomez22ral}. 
Moreover, ERL-V aims at reducing the complexity of the camera ego-motion problem in \cite{Jaegle16icra} by assuming zero angular velocity (\cref{sec:method}). 
To validate this benefit, we measure the runtime performance of \mbox{ERL-V} and ERL when processing stabilized and non-stabilized data, respectively. 
For this purpose, we randomly selected pairs of consecutive images (grayscale frames or IWEs) to determine the processing time for estimating the camera's velocity. 
For each trial, 500 optical flow vectors were used to estimate the ego-motion. 
ERL and ERL-V provide linear velocity estimations in \SI{0.72}{\second} and \SI{0.53}{\second} using Matlab. All time processing experiments ran in an Intel\textregistered i7-7700HQ CPU, which could be further reduced by using specialized hardware. 
The results indicate that stabilization provides an average processing time reduction of 25.78\% (ERL-V) with respect to non-stabilized (ERL). 

\vspace{-1.5ex}
\subsection{Feature tracking evaluation}
\label{sec:experim:featuretracking}

Next, we compare the performance of state-of-the-art tracking algorithms by processing stabilized and non-stabilized data. 
For this evaluation, we select the well-known KLT algorithm to perform frame-based feature tracking, and EKLT \cite{Gehrig19ijcv} for event-based tracking. 
Despite the learning approach in \cite{Messikommer23cvpr} performs better than EKLT, we prefer the latter as it provides the best tracking results among traditional approaches and allows us to analyze methods without dependency on learning-based stages. 
A more complex validation with learning-based stabilization and baselines is left as future work.

\subsubsection{\textbf{Ground truth feature tracks}}
We compute GT feature tracks (i.e., references) using the image alignment algorithm in \cite{Evangelidis2008tpami}, which estimates the motion model between pairs of frames using the Enhanced Correlation Coefficient (ECC). 
Due to how ECC is applied (on homographic warps), the evaluation considers only sequences recorded on planar scenes, such as \emph{boxes 6DOF}, \emph{boxes rotation}, \emph{poster 6DOF}, and \emph{poster rotation} from \cite{Mueggler17ijrr}. 
For each experiment, we extract an initial set of frame-based features $F_\text{init}$ using the Harris corner detector \cite{Harris88}. 
$F_\text{init}$ initializes the GT and the feature tracking methods (i.e., KLT and EKLT). 
Afterwards, GT references are propagated using the warp transformations provided by ECC for each pair of consecutive frames. 
On the other hand, stabilized GT references are obtained by applying the orientation-stabilized transformation $\Rot_{k}$ (see \cref{eq:event_stabilization}) to each set of track references $F_{k}$. 
Finally, if a saccadic motion occurs due to stabilization, the GT references and the tracking algorithms are restarted. 

\subsubsection{\textbf{Evaluation metrics}}
Inspired by \cite{Gehrig19ijcv,Messikommer23cvpr}, we adopt the following performance metrics: Tracking Error (\emph{TE}), Track-normalized  Error (\emph{TTE}), End-point Tracking Error (\emph{ETE}), Normalized Feature Age (\emph{NFA}), and Track-normalized Feature Age (\emph{TFA}). 
Tracking error corresponds to the average Euclidean distance between valid-track features and their GT, where valid-track refers to features tracked for at least $\tau$ seconds. 
For each experiment, $\tau$ is set to the inverse of the APS frame rate of the evaluation dataset. 
Similarly, \emph{TTE} pertains to the tracking errors normalized by the total number of tracking samples, while \emph{ETE} is the Euclidean distance between the last tracked sample and its GT. 
Further, \emph{NFA} corresponds to the average valid-track feature ages normalized by the age of their GT, while \emph{TFA} refers to normalized feature ages divided by the number of tracking samples.

\subsubsection{\textbf{Processing pipeline}}
The stabilization pipeline follows the same configuration as in \cref{sec:linearvelestination} except for the event \emph{windowing} approach. 
In these experiments, event \emph{windowing} is synchronized with the timestamp of the grayscale images so that frames and events are stabilized with the same camera orientation. 
The spatio-temporal synchronization is necessary to match the ECC GT information with the event tracking feature provided by the baseline EKLT.

\subsubsection{\textbf{Results. Feature tracking using frames}}
\Cref{tab:experim:tracking-all} summarizes the frame-based feature tracking results. 
In almost all cases, the tracking error metrics (\emph{TE}, \emph{TTE}, and \emph{ETE}) obtained with stabilized frames outperform the non-stabilized tracking ones. 
In general, stabilization improves \emph{TE} by 33.6\%, \emph{TTE} by 33.1\%, and \emph{ETE} by 34.9\%, which highlights the benefits of stabilization in terms of tracking accuracy. 
The performance improvements are mainly due to the simplification of the camera motion through stabilization, where track features describe almost translational motions. 
These benefits are more evident in rotational sequences, where features from stabilized frames describe only the residual linear displacements of the camera.
Moreover, stabilization marginally affects feature tracking age:
while tracking with non-stabilized frames achieves better values, the percentage of improvement is markedly smaller (1.6\% in \emph{FA} and 3.7\% in \emph{TFA}) than the accuracy gains. 
The tracking age performance differences are mainly due to the frame warping when stabilizing for large changes of orientation. 
Under this condition, the feature appearance between stabilized frames might vary, which affects KLT performance. 
It is worth noting that this behavior is reported very few times in our experiments and usually precedes a system saccade.

\subsubsection{\textbf{Results. Event-based feature tracking}}
To measure the tracking error using events we follow the approach in \cite{Gehrig19ijcv}, which compares the GT feature tracks (obtained via ECC) with the estimated ones by linearly interpolating the two closest event-based feature track locations in time. 
The tracking error corresponds to the distance between GT and the position of the interpolated feature track.
Results are presented in \cref{tab:experim:tracking-all}. 
In most cases, the metrics obtained with event stabilization are better than those with non-stabilized data. 
Our stabilization approach improves the \emph{TTE} and \emph{ETE} performance by 36.2\% and 28.2\%, while providing similar \emph{TE} results to the non-stabilized approach.
In terms of feature age, non-stabilized event tracking improves by \emph{FA} (5.68 \%) and \emph{TFA} (4.82 \%), which are still small compared to the tracking accuracy gains.
Feature age performance decreases when tracking features with large spatial displacements (i.e., events stabilized for large changes of orientation). 
In this case, the quantization of the event coordinates after stabilization modifies their spatial distribution, which might cause feature loss when processing patches of stabilized events.

\vspace{-1.5ex}
\section{Conclusion}
\label{sec:conclusion}

We have presented a method to visually stabilize the changes of camera orientation.
To the best of our knowledge, this is the first work that studies the benefits of stabilization by comparing the performance of frame- and event-based methods while processing stabilized and non-stabilized data. 
The experimental results demonstrate that stabilization increases feature tracking and camera velocity estimation accuracy by 27.37\% and 34.82\%, respectively. 
Besides, it decreases the processing time of camera ego-motion estimation by at least 25\% by reducing the complexity of the problem. 

The experimental results confirm the advantages of integrating visual stabilization in robotics. 
Nonetheless, incorporating stabilization into perception pipelines presents some unique considerations. 
First, stabilization requires accurate camera orientation estimation to guarantee the performance reported in \cref{sec:experim}. 
A simple solution consists of estimating the camera orientation from the measurements provided by a high-quality IMU. 
Nowadays robot platforms include either cameras with embedded internal sensors or external IMUs that directly provide the sensor orientation \cite{RodriguezGomez22ral,Gao22ral}, which makes stabilization a reality. 
Second, the rate of the camera orientation provider must be faster than the camera frame rate or the duration of the event slice / window. 
Otherwise, stabilization might be undermined as there is a large temporal difference between the visual data and the attitude estimation. 
Finally, if the camera motion describes small orientation changes, stabilization might be unimportant, as the geometric action of compensation has a negligible effect on the visual data.
Therefore, as expected, the benefits of stabilization are most noticeable in applications where the camera suffers from regular or aggressive changes of orientation (e.g., data collected onboard drones). 
\vspace{-1.5ex}

\bibliographystyle{IEEEtran}

\end{document}